\documentclass[10pt,twocolumn,letterpaper]{article}

\usepackage{cvpr}
\usepackage{times}
\usepackage{epsfig}
\usepackage{graphicx}
\usepackage{amsmath}
\usepackage{amssymb}
\usepackage{enumerate}
\usepackage{booktabs}
\usepackage{enumitem}

\usepackage[breaklinks=true,bookmarks=false]{hyperref}

\cvprfinalcopy %

\def\cvprPaperID{1134} %

\begin{document}

\title{Spatio-temporal Video Re-localization by Warp LSTM}

\author{Yang Feng$^\sharp$\thanks{This work was done while Yang Feng was a Research Intern with Tencent AI Lab.} \qquad Lin Ma$^{\natural}$\thanks{Corresponding author.} \qquad Wei Liu$^\natural$ \qquad Jiebo Luo$^\sharp$\\
$^\natural$Tencent AI Lab \quad $^\sharp$University of Rochester\\
{\texttt{\small \{yfeng23,jluo\}@cs.rochester.edu\qquad forest.linma@gmail.com \qquad wl2223@columbia.edu}}
}

\maketitle

\begin{abstract}
The need for efficiently finding the video content a user wants is increasing because of the erupting of user-generated videos on the Web. Existing keyword-based or content-based video retrieval methods usually determine what occurs in a video but not when and where. In this paper, we make an answer to the question of when and where by formulating a new task, namely spatio-temporal video re-localization. Specifically, given a query video and a reference video, spatio-temporal video re-localization aims to localize tubelets in the reference video such that the tubelets semantically correspond to the query. To accurately localize the desired tubelets in the reference video, we propose a novel warp LSTM network, which propagates the spatio-temporal information for a long period and thereby captures the corresponding long-term dependencies. Another issue for spatio-temporal video re-localization is the lack of properly labeled video datasets. Therefore, we reorganize the videos in the AVA dataset to form a new dataset for spatio-temporal video re-localization research. Extensive experimental results show that the proposed model achieves superior performances over the designed baselines on the spatio-temporal video re-localization task.
\end{abstract}

\section{Introduction}

Video sharing websites or APPs are becoming more popular than ever before. Besides the traditional video-sharing websites, including YouTube\footnote{\url{https://www.youtube.com}} and Facebook\footnote{\url{https://www.facebook.com}}, the recently emerged short video sharing APPs, such as SnapChat\footnote{\url{https://www.snapchat.com}} and TikTok\footnote{\url{https://www.tiktok.com}}, arouse the passion of ordinary users for creating and sharing video contents. With more and more videos generated every day, exploring so many videos becomes increasingly challenging. It is necessary to build tools which can help users find the video contents they want efficiently.

\begin{figure}
  \centering
  \includegraphics[width=\columnwidth]{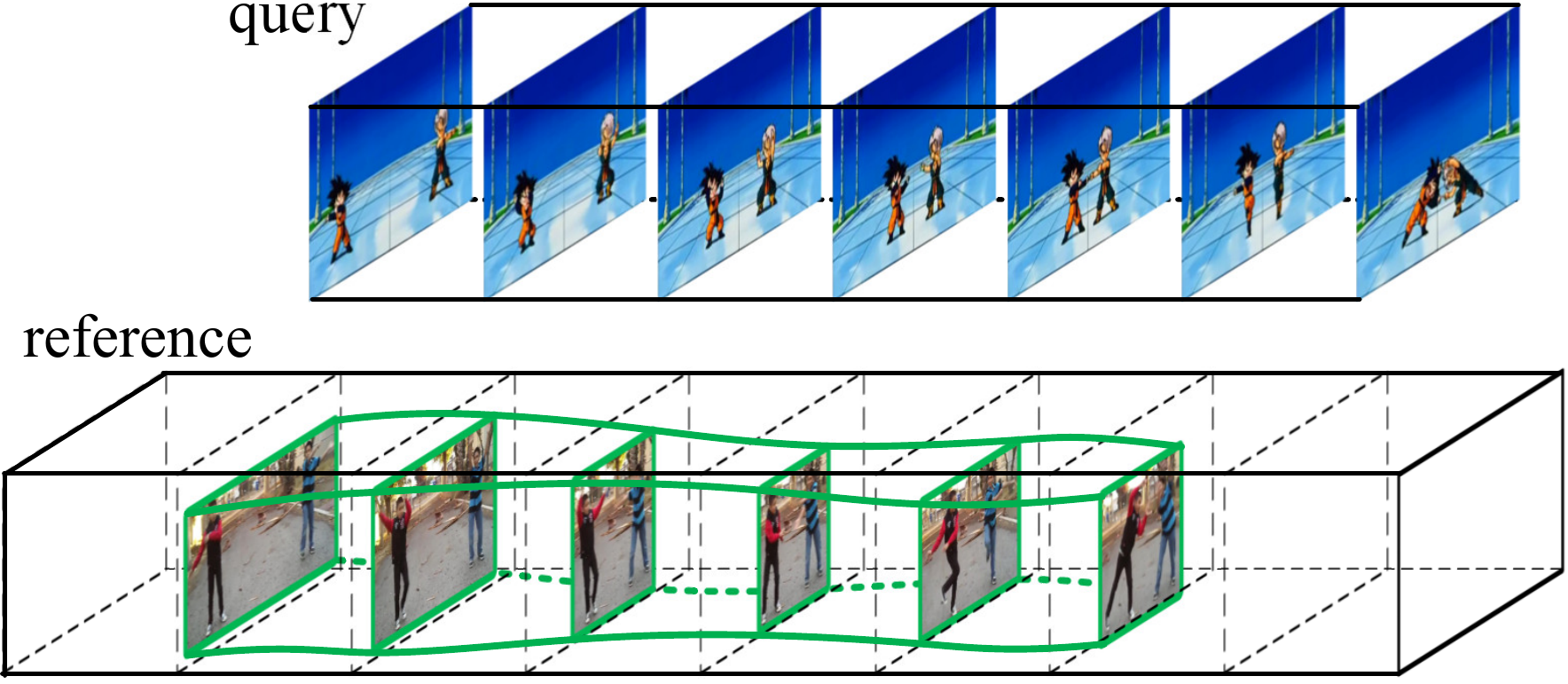}
  \caption{The query is a video containing an action performed by two characters. The reference video contains two boys performing the same action. 
  Given the query, spatio-temporal video re-localization aims to localize the tubelets in the reference video such that the tubelets express the same visual concept as the query. The desired tubelet in the reference is marked by green. Best viewed in color.}
  \label{fig:1}
\end{figure}

Keyword-based video search is prevalent among users when they want to find some videos. %
Although it is a powerful method, keyword-based video search results are largely determined by the text information associated with the videos. %
As such, content-based video retrieval (CBVR) methods \cite{cao2013mining,cao2015localizing,chang1998fully,chen2009human,hsieh2006motion,jiang2016hierarchical,su2007motion,visser2002object,yan2003negative} are proposed for tackling this problem. %
With the indexing and retrieval techniques, a large list of videos is returned by the CBVR system. Only the top results will be viewed by a user, as it is time-consuming to browse the whole video from the beginning to the end and thereby determine the relevance.

Two kinds of methods are designed to avoid browsing the whole video. The first kind is video summarization methods~\cite{kanehira2018aware,zhao2018hsa}, which generate a short synopsis for a long video. The second kind of methods~\cite{chen2018temporally,chen2019localizing,feng2018video,gao2017tall,hendricks2017localizing,hou2017tube,kalogeiton2017action,li2018recurrent,shou2018online} try to trim the video segment of interest. Using natural language as a query, \cite{gao2017tall,hendricks2017localizing} retrieve a specific temporal segment in a video, which shares the same semantic meaning as the query. By replacing the query sentence with a sample video clip, video re-localization~\cite{feng2018video} aims to temporally localize video segments, which semantically correspond to the query video clip. %

In this paper, we extend the temporal video re-localization~\cite{feng2018video} to the spatio-temporal domain. Specifically, given a query video, spatio-temporal video re-localization~(STVR) aims to localize tubelets in a reference video such that the tubelets are semantically coherent with the query video. Figure~\ref{fig:1} illustrates an example pair of query and reference videos. There are several advantages in localizing tubelets over temporal localization based on whole frames. First, localizing tubelets can handle the cases where multiple events are happening at the same time in the reference video. When using whole video frames for recognition or temporal detection tasks, it usually assumes that only one event is undergoing. The assumption rarely holds in unconstrained environments. Second, the recognition accuracy will substantially increase because the influence of background regions is reduced by only focusing on specific regions. STVR is also a more challenging task than temporal video re-localization. First, the training videos should be labeled with bounding boxes over a long period, which consumes more human labors than temporal annotation only. Second, detecting the bounding boxes at each frame is more difficult than only localizing the starting and ending boundary points.

To address the STVR task, we propose a matching framework consisting of three modules: query encoding, reference encoding, and query-reference interaction. The query encoding module encodes a query video of arbitrary resolution into a series of fixed size feature cubes. The reference encoding module encodes the given reference video in a different manner. To keep the detailed spatio information in the reference, the shape of the reference feature cube is proportional to the resolution of the reference video. In the query-reference interaction module, several bounding box proposals are generated for the reference and then each proposal is matched with the query to determine whether a proposal and the query are semantically corresponding to each other.

To accurately localize the tubelets in the reference, the long-term spatio-temporal information needs to be modeled. We propose a novel warp LSTM network for this purpose. Warp LSTM is a variant of ConvLSTM \cite{xingjian2015convolutional}. In ConvLSTM, the previous hidden state is concatenated with the current input for further computation. Different from ConvLSTM, the previous hidden state in warp LSTM is warped before the concatenation to make the previous hidden state be aligned with the current input if any movement in the video makes them unaligned. The warp of the hidden state at a previous time-step can compensate for small movements in the video, which accurately aggregates the spatio history information of moving objects.

In order to train the matching model for STVR, we create a new dataset by reorganizing the videos in the AVA dataset \cite{gu2018ava}. The AVA dataset is originally used for spatio-temporal action localization. Each action tubelet is annotated with one or several atomic action labels. We use one action tubelet as the query and find the tubelets with the same action labels in the reference video. Two action tubelets are semantically corresponding to each other if the action labels of the two tubelets are exactly the same. The AVA dataset provides a subset of videos for training and another subset of videos for validation. We further split the action tubelets into training, validation, and test subsets according to their action categories. Such a splitting guarantees that the validation and testing categories do not overlap with the training categories.

In summary, our contributions are four-fold:
\setlist{nolistsep}
\begin{itemize}[noitemsep]
	\item We make the first attempt to tackle the STVR task, which aims to localize tubelets in the reference video such that the tubelets semantically correspond to a given query video.
    \item We propose a novel warp LSTM network to propagate the spatio-temporal information between adjacent frames for a long period and thereby capture the corresponding long-term dependencies.
    \item We reorganize the videos in the AVA dataset \cite{gu2018ava} to form a new dataset for the research on STVR. 
    \item We conduct extensive experiments on the new dataset, which shows that the warp LSTM performs better than the competing methods.
\end{itemize}

\section{Related Work}
\textbf{Video Representations.} Convolutional Neural Networks (CNNs) have broken many records of computer vision tasks, such as image classification~\cite{he2016deep,krizhevsky2012imagenet}, object detection~\cite{huang2017speed}, semantic segmentation~\cite{chen2018encoder}, facial expression recognition \cite{zhang2017facial}, and captioning~\cite{chen2018regularizing,jiang2018recurrent,wang2018reconstruction,wang2018bidirectional,you2016image}. Due to the great success of CNNs on images, many researchers tried to apply CNNs on videos. %
\cite{karpathy2014large,simonyan2014two,yue2015beyond} are mainly based on 2D CNNs, in which the motion information is not fully exploited. 3D CNNs are proposed in \cite{ji20133d,tran2015learning,zhang2019adversarial} to capture more complex motion patterns. The recently proposed I3D feature \cite{carreira2017quo} has achieved state-of-the-art action recognition results. Compared with 3D CNNs, the proposed warp LSTM is able to model the long-term spatio-temporal information of moving objects for classification and localization tasks by explicitly modeling the movements in videos.

\textbf{Video Re-localization.} Video Re-localization \cite{feng2018video} aims to find segments in reference videos semantically corresponding to a given query video. %
A more specialized task, one-shot action localization \cite{yang2018one}, focuses on the temporal detection of actions in videos giving an example. The STVR task to be solved in this paper is an extension of temporal video re-localization. Besides predicting the starting and ending points of a video segment, STVR also detects the spatio localization of the video content that users are interested in. Hoogs \textit{et al.} \cite{hoogs2015end} designed a system to spatio-temporally retrieve people and vehicles in surveillance videos. STVR is different in that it is not specialized in certain categories or types of videos.

\begin{figure}
  \centering
  \includegraphics[width=\columnwidth]{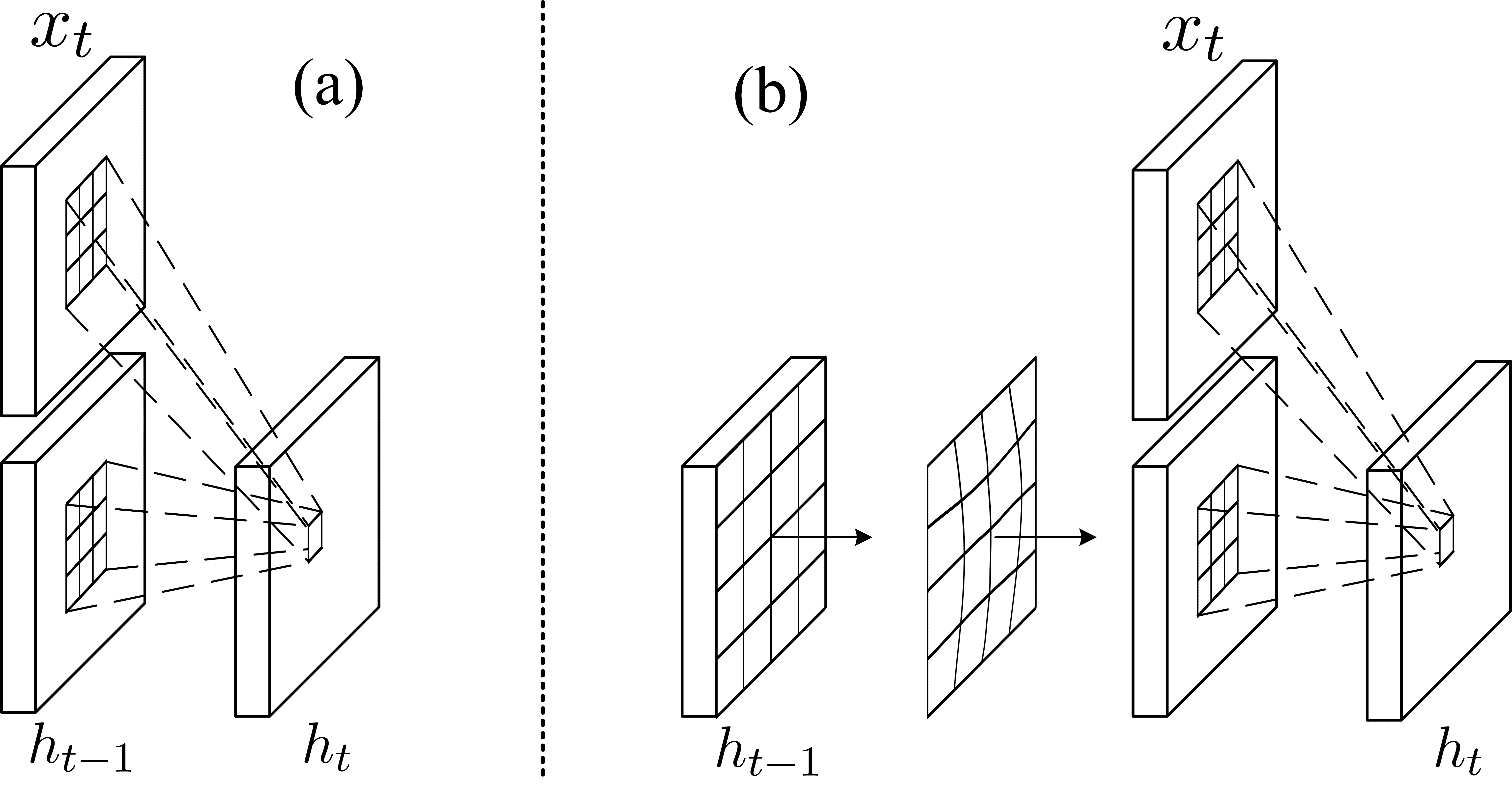}
  \caption{Comparison between ConvLSTM and warp LSTM. (a) In ConvLSTM, the input $x_t$ and the hidden state at previous time-step $h_{t-1}$ are convolved with different filters and then added to produce the new hidden state $h_t$. (b) In warp LSTM, $h_{t-1}$ is warped by a differentiable spline interpolation before the convolution to compensate for the small motion between consecutive video clips.}
  \label{fig:warp}
\end{figure}

\textbf{Spatio-temporal Detection.} Two related vision tasks are video object detection and spatio-temporal action detection. All the three tasks need to model long-term spatio-temporal dependencies to predict tubelets in videos. %
Although the temporal information inside a clip is considered in \cite{hou2017tube,kalogeiton2017action}, the bounding boxes are predicted independently of the frames outside the short clip. To solve this problem, both~\cite{kang2017object,li2018recurrent} extract tubelet proposals from videos in the first stage and make the classification in the second stage. %
One assumption used in both \cite{kang2017object,li2018recurrent} is that the reception field of CNN features is large enough to handle the small movements in a short time. With this assumption, the feature cropped at a previous anchor location is used to predict the bounding boxes at the current frame. Although the reception field is large enough to cover the objects with small movements, the bounding box prediction will become a more difficult task on the feature map with offsets. Different from them, we align the previous feature maps with the current feature map by warping. The proposed warp LSTM can reduce the offset of the previous feature map and thereby reduce the burden of the bounding box prediction module.

\section{Spatio-temporal Information Propagation}
In this section, we present our proposed warp LSTM network for modeling the long-term spatio-temporal information in videos. Warp LSTM is a variant of ConvLSTM~\cite{xingjian2015convolutional}. We first give the background knowledge of ConvLSTM.
\subsection{ConvLSTM}
ConvLSTM extends the fully-connected LSTM \cite{hochreiter1997long} to have convolutional structures in both the input-to-state and state-to-state transitions:
\begin{equation}
\label{eq:convlstm}
\begin{aligned}
i_t &= \sigma(W_{xi}*x_t+W_{hi}*h_{t-1}+b_i), \\
g_t &=  \sigma(W_{xg}*x_t+W_{hg}*h_{t-1}+b_g), \\
f_t &= \sigma(W_{xf}*x_t+W_{hf}*h_{t-1}+b_f), \\
o_t &= \sigma(W_{xo}*x_t+W_{ho}*h_{t-1}+b_o), \\
c_t &= f_t \odot c_{t-1} + i_t \odot g_t, \\
h_t &= o_t\odot \phi(c_t),
\end{aligned}
\end{equation}
where $x_t$, $h_t$, $c_t$, $i_t$, $f_t$, and $o_t$ are the ConvLSTM input, hidden state, memory cell, input gate, forget gate, and output gate at time-step t, respectively. All the $W$s and $b$s are the parameters of the ConvLSTM layer. $*$ is the convolution operation and $\odot$ is the element-wise product. $\sigma$ and $\phi$ are sigmoid non-linearity and hyperbolic tangent nonlinearity, respectively. In Eq. (\ref{eq:convlstm}), $x_t$ and $h_{t-1}$ are convolved with different filters and then added for later computation, as shown in Figure~\ref{fig:warp}. This operation is equivalent to first concatenating $x_t$ and $h_{t-1}$ along the channel dimension and then computing the convolution.

\begin{figure}
  \centering
  \includegraphics[width=3cm]{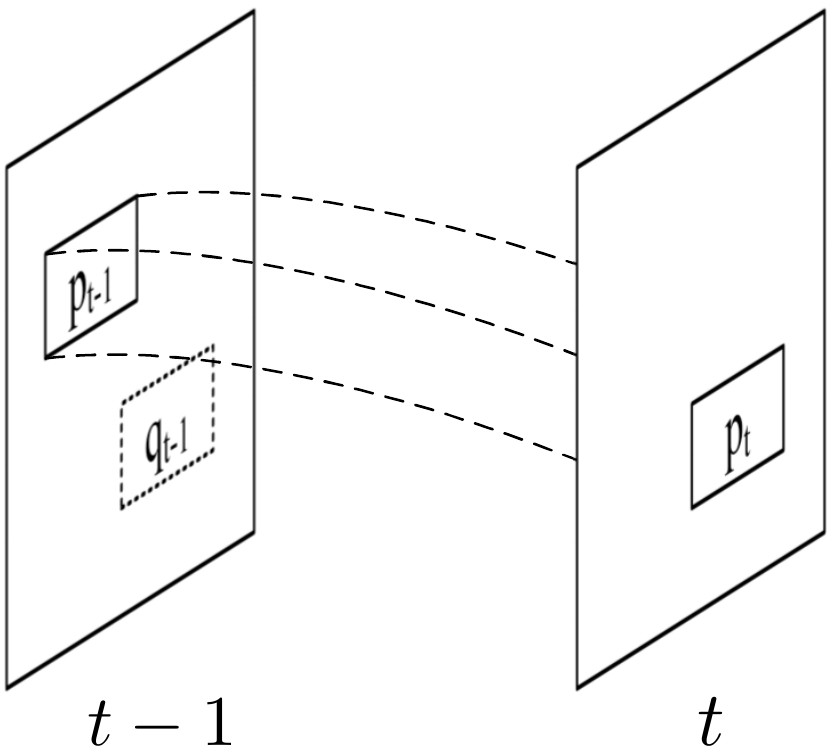}
  \caption{An illustration of a moving object in two consecutive time-steps. At time-step $t-1$, the object is located at the bounding box $p_{t-1}$. In the following time-step, the object moves to the location of bounding box $p_t$. $q_{t-1}$ is a bounding box at time-step $t-1$ having the same position as $p_t$. Please note that the content in $q_{t-1}$ may be not semantically related to the object in $p_{t}$.}
  \label{fig:move}
\end{figure}

\subsection{Warp LSTM}
If no movement happens in the video at the $t$-th time-step, the concatenation of $x_t$ and $h_{t-1}$ in ConvLSTM is perfectly fine. %
However, when motion happens at time-step $t$, the concatenation of $x_t$ and $h_{t-1}$ may cause errors in spatio-temporal localization tasks. Figure~\ref{fig:move} shows a moving object at two consecutive time-steps. At time-step $t-1$, the object is located at the bounding box $p_{t-1}$. In the following time-step, the object moves to the location of bounding box $p_t$. $q_{t-1}$ is a bounding box at time-step $t-1$ having the same position as $p_t$. The content in $q_{t-1}$ may be depicting objects other than the aforementioned object. As such, simply concatenating the features at the locations of $q_{t-1}$ and $p_t$ may introduce noises into the classification and localization of $p_t$.

\begin{figure}
  \centering
  \includegraphics[width=\columnwidth]{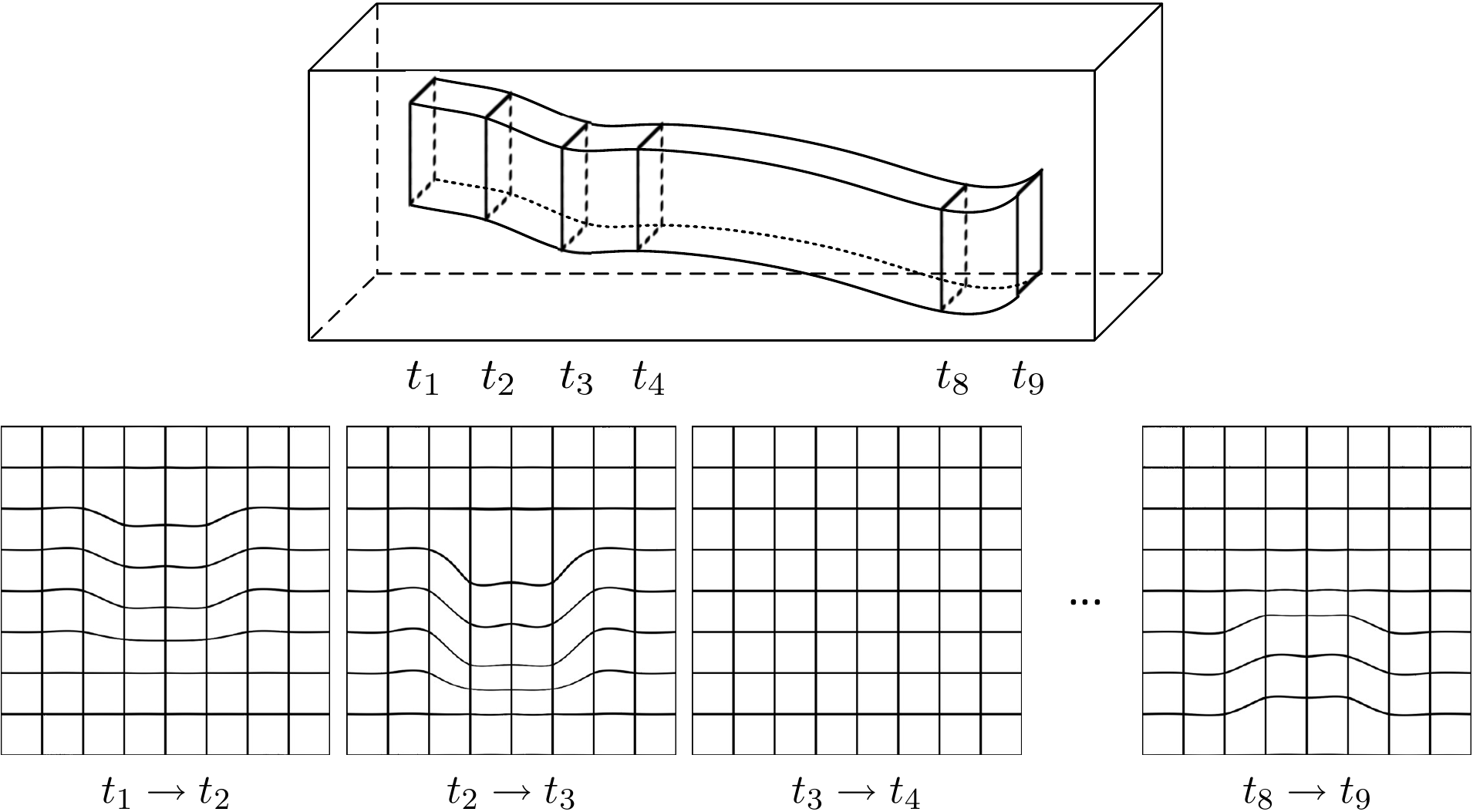}
  \caption{The illustration of the warp obtained by polyharmonic interpolation. It can be observed that the spatio-temporal information of a moving object is propagated for a long period.}
  \label{fig:warp_illu}
\end{figure}

We propose warp LSTM to address this issue by warping the hidden state at the previous time-step before concatenating it with the input. Figure~\ref{fig:warp}  illustrates the proposed warp LSTM, where the warp is implemented by the differentiable spline interpolation~\cite{cole2017synthesizing}, %
as illustrated in Figure~\ref{fig:warp_illu}. Given a set of 2-D control points $\{(x_1,y_1),\ldots,(x_n,y_n)\}$ on the hidden state $h$, the warp operation tries to shift $(x_i,y_i)$ to a new position $(x_i+dx_i, y_i+dy_i)$, where $n$ is the number of control points and $(dx_i,dy_i)$ is the desired displacement of the $i$-th control point. Let $h'$ denote the warped hidden state, then we have:
\begin{equation}
h'[x_i+dx_i, y_i+dy_i]=h[x_i,y_i], \forall i\in \{1,\ldots,n\}.
\end{equation}
Besides shifting the control points, the warping is continuous on the whole 2D space of $h$, resulting in a dense flow field. The flow field is estimated by the polyharmonic interpolation~\cite{iske2004multiresolution}:
\begin{equation}
\label{eq:interp}
s(x,y)=\sum_{i=1}^nw_i\phi_k(\|(x,y)-(x_i,y_i)\|)+v_1x+v_2y+v_3,
\end{equation}
where $\phi_k$ is a set of radial basis functions. $w_i$, $v_1$, $v_2$, and $v_3$ are interpolation parameters. After optimization, the polyharmonic interpolation $s$ will shift the control points exactly to their desired locations. In addition, the warped $h'$ is a differentiable function of $h$, $(x_i,y_i)$, and $(dx_i,dy_i)$.

In practice, the control points are fixed in advance. We evenly put horizontal lines and vertical lines in the 2D space of $h$ and put control points on the intersections of horizontal and vertical lines. The displacement $(dx_i, dy_i)$ is predicted by an additional convolutional layer. We also add extra control points with zero displacements at the boundary. Two radial basis functions, \textit{i.e.}, $\phi_1(r)=r$ and $\phi_2(r)=r^2\log(r)$, are chosen for the interpolation.
The proposed warp LSTM is defined as:
\begin{equation}
\label{eq:warplstm}
\begin{aligned}
d_{t-1} &= W_{xd}*x_t+W_{hd}*h_{t-1}+b_d, \\
h_{t-1}' &= \text{warp}(h_{t-1}, d_{t-1}), \\
c_{t-1}' &= \text{warp}(c_{t-1}, d_{t-1}), \\
i_t &= \sigma(W_{xi}*x_t+W_{hi}*h_{t-1}'+b_i), \\
g_t &=  \sigma(W_{xg}*x_t+W_{hg}*h_{t-1}'+b_g), \\
f_t &= \sigma(W_{xf}*x_t+W_{hf}*h_{t-1}'+b_f), \\
o_t &= \sigma(W_{xo}*x_t+W_{ho}*h_{t-1}'+b_o), \\
c_t &= f_t \odot c_{t-1}' + i_t \odot g_t, \\
h_t &= o_t\odot \phi(c_t),
\end{aligned}
\end{equation}
where $d_{t-1}$, $h_{t-1}'$, and $c_{t-1}'$ are the displacement, warped hidden state, and warped memory cell at time-step $t-1$, respectively. %
$\text{warp}(\cdot,\cdot)$ is the sparse image warping function\footnote{\url{https://www.tensorflow.org/api_docs/python/tf/contrib/image/sparse_image_warp}}, which warps an image based on control point displacements.

\textbf{Discussion.} The closest work to our proposed warp LSTM is TrajGRU~\cite{shi2017deep}, which also warps the feature map at a previous time-step to the current time-step. However, there are two major differences between the two methods. The motivation of TrajGRU is to learn a dynamic connection structure, \textit{e.g.}, replacing the fixed $3\times3$ convolution with 5 learned dynamic links. Our motivation is to align the previous feature map with the current feature map. Several dense flow fields are predicted by convolutional layers in TrajGRU for warping, while the displacements of a set of control points are predicted in the warp LSTM. The warp computed by polyharmonic interpolation is continuous everywhere, while the dense flows generated by convolutional layers in TrajGRU may be not. TrajMF~\cite{jiang2015human} is also designed for explicit motion modeling. Compared with TrajMF, the proposed warp LSTM is not handcrafted and is thus benefiting from the feature learning ability of deep neural networks.

\begin{figure*}
  \centering
  \includegraphics[width=\textwidth]{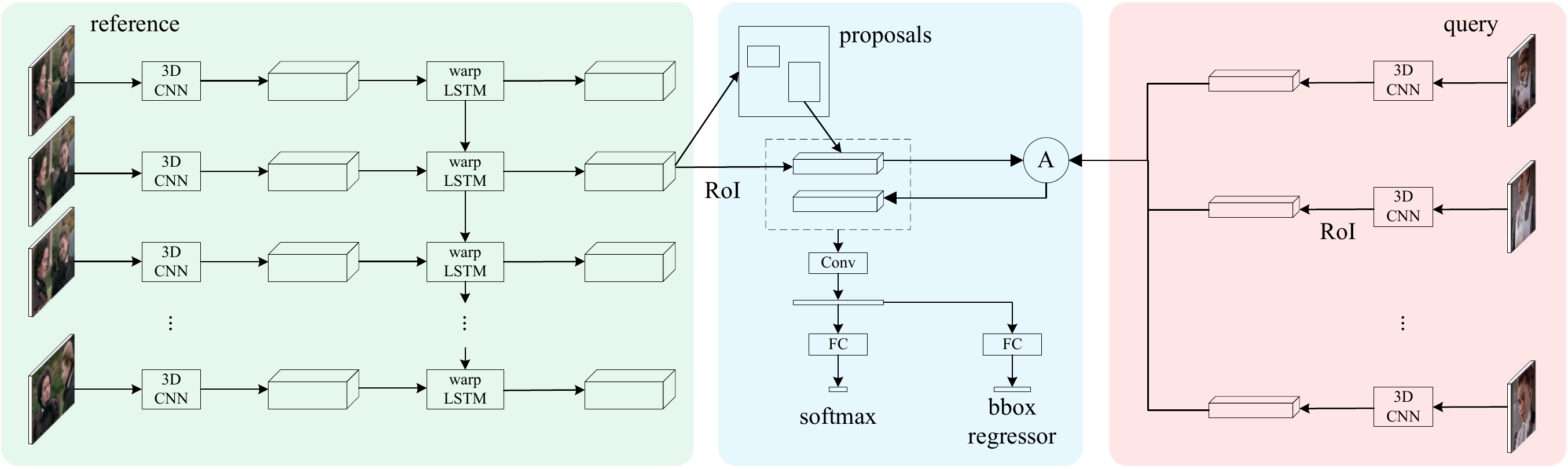}
  \caption{The architecture of our proposed model for STVR. The inputs are a query and a reference video. Both query and reference are split into clips and then fed into a 3D CNN to extract video features. Later, the long-term spatio-temporal information in the reference video is aggregated by the warp LSTM to produce a new reference feature. Region proposal network~\cite{ren2015faster} is applied on the new reference feature to generate several proposals. For each proposal, we use an attention mechanism to select the most related query feature and concatenate the proposal feature with the attention weighted query feature. The concatenated feature is used for the second stage prediction, which outputs a refined bounding box and a binary label indicating whether the query and the proposal are semantically corresponding to each other. \textcircled{{\tiny\fontfamily{phv}\selectfont A}} denotes the attention mechanism, and the dashed rectangle means concatenating along the channel dimension.}
  \label{fig:framework}
\end{figure*}

\section{Spatio-temporal Video Re-localization}
Given a query video and a reference video, STVR aims to localize tubelets in the reference video such that the tubelets semantically correspond to the query video. To achieve the goal, we design a novel model detecting bounding boxes in the reference video based on the matching results between the query and reference videos. Our proposed model is shown in Figure \ref{fig:framework}.

\subsection{Video Feature Extraction}
For the STVR task, both the temporal and spatio information should be captured in the raw video feature. Hence, we choose inflated 3D ConvNet (I3D)~\cite{carreira2017quo} as the feature extractor. The I3D model is originally trained on 64-frame video snippets and tested on 250-frame video snippets. Using many frames together to extract video features is fine for video classification task, but it may not be a good idea for the spatio-temporal localization task because the regions we want to locate may move over a long distance. Therefore, we reduce the number of frames of the video snippets to 8. We also re-sample all the videos at the FPS of 24 so that each snippet is just $\frac{1}{3}$ second long. We choose the activation values at the ``Mixed\_4c'' layer in the I3D model as the video feature, which has a spatio stride of 16 and a temporal stride of 4.

Let $r_i\in\mathbb{R}^{8\times H\times W\times3}$ denote the $i$-th reference clip, where $H$ and $W$ are the height and width of the reference video, respectively. The feature extraction for the reference is given by:
\begin{equation}
\hat{f}^r_i=\text{I3D}(r_i),
\end{equation}
where $\hat{f}^r_i\in\mathbb{R}^{2\times\frac{H}{16}\times\frac{W}{16}\times512}$ is the extracted feature for the $i$-th reference clip. The 4D feature is transformed to 3D by flattening along the temporal dimension and the channel dimension:
\begin{equation}
f^r_i = \text{flatten}(\hat{f}^r_i),
\end{equation}
where $f^r_i\in\mathbb{R}^{\frac{H}{16}\times\frac{W}{16}\times1024}$ is the flattened feature. 
For the $j$-th clip in the query video $q_j$, we apply 2D RoI pooling after 3D convolution to generate a fixed size feature:
\begin{equation}
f^q_j=\text{RoI}\left(\text{flatten}\big(\text{I3D}(q_j)\big)\right),
\end{equation}
where $f^q_j\in\mathbb{R}^{7\times7\times1024}$ is the $j$-th query feature.

\subsection{Reference Propagation}
The extracted $f_i^r$ only contains the spatio-temporal information within the 8-frame clip. To propagate the spatio-temporal information from previous clips of the reference video to the $i$-th clip for better re-localization, we add a warp LSTM layer to update the reference feature.
\begin{equation}
h_i=\text{warpLSTM}(f_i^r,h_{i-1}),
\end{equation}
where $h_i\in\mathbb{R}^{\frac{H}{16}\times\frac{W}{16}\times1024}$ is the hidden state of the warp LSTM, which also serves as a new reference representation.

\subsection{Proposal Generation}
The proposal generation module aims to find all the bounding boxes containing the content of potential interest in one clip. The generation of reference proposals is designed following Faster RCNN~\cite{ren2015faster}. $h_i$ is fed into the region proposal network (RPN) to generate proposals:
\begin{equation}
\begin{aligned}
p_k &= \text{RPN}(h_i), \\
f^p_k &= \text{RoI}(h_i, p_k),
\end{aligned}
\end{equation}
where $p_k$ is the predicted bounding box for the $k$-th proposal and $f^p_k\in\mathbb{R}^{7\times7\times1024}$ is the feature of the $k$-th proposal obtained by RoI pooling.

\subsection{Query and Reference Matching}
We match every proposal in the reference clip with the query video. The query video may be much longer than one clip in the reference, which has only 8 frames. 
As such, some parts in the query video may not well correspond to a short proposal. Motivated by~\cite{feng2018video,wang2016machine}, we design an attention mechanism to select which part in the query video should be matched with the proposal. For the $k$-th proposal, the features of the query video are attentively summarized as:
\begin{equation}
\label{eq:att}
\begin{aligned}
e_{k,j} &= \tanh(W^q\text{avg}(f^q_j) + W^r\text{avg}(f^p_k) + b_p), \\
\alpha_{k,j} &= \frac{\exp(w^\top e_{k,j} + b_s)}{\sum_i \exp(w^\top e_{k,i} + b_s)}, \\
\bar{f}^q_k &= \sum_j \alpha_{k,j}f^q_j,
\end{aligned}
\end{equation}
where $\bar{f}^q_k$ is the weighted query representation. $W^q$, $W^r$, $w$ are the weight parameters in the attention model with $b_p$ and $b_s$ denoting the bias terms. %
$\text{avg}(\cdot)$ means average pooling along the spatio dimensions.

\subsection{Label and Bounding Box Predictions}
The proposal feature $f^p_k$ and the attentively weighted query feature $\bar{f}^q_k$ are concatenated along the channel dimension for the final label prediction and bounding box refinement. The final label is binary, indicating whether the query video and the $k$-th proposal are semantically corresponding to each. The ground-truth label will be ``true'' if the query video and the $k$-th proposal are indeed semantically corresponding to each other. Otherwise, the ground-truth label will be ``false''.
The bounding box regression layers are designed following \cite{ren2015faster}. Please refer to \cite{ren2015faster} for more details.

\begin{figure}
\vspace{-0.4cm}
  \centering
  \includegraphics[width=0.9\columnwidth]{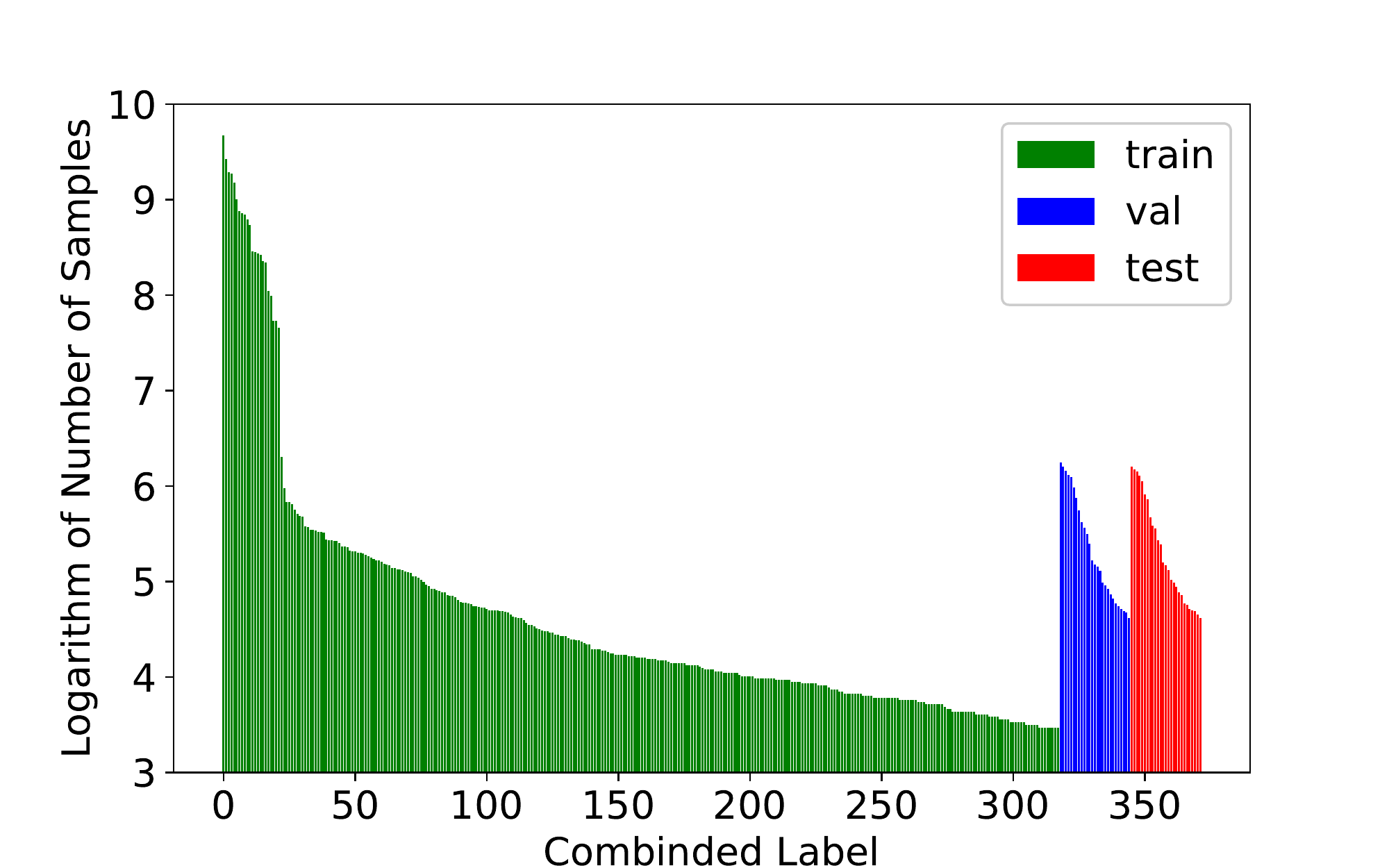}
  \caption{The distribution of the number of samples for each combined label. The combined labels belonging to training, validation, and testing sets are marked by green, blue, and red, respectively. The combined labels with less than 32 tubelet samples in the training set are omitted for clarity.}
  \label{fig:hist}
\vspace{-0.4cm}
\end{figure}

\section{The Reorganized Datasets}
\label{sec:ava}
Existing video datasets are designed for other vision tasks, such as classification~\cite{kay2017kinetics}, temporal localization~\cite{caba2015activitynet}, action recognition~\cite{soomro2012ucf101}, captioning~\cite{chen2011collecting}, and video summarization~\cite{gygli2014creating}. None of them is suitable for the STVR task, which requires pairs of query and reference videos. The query should semantically correspond to  some labeled tubelets in the reference video. It will require a huge expensive labor  to collect and annotate such a video dataset.

As such, we propose to reorganize the AVA dataset for the STVR task. The AVA dataset~\cite{gu2018ava} is originally designed for the spatio-temporal action localization task. There are 430 15-minute video clips with per second action bounding box annotations. The annotated actions are 80 categories of atomic actions, including ``stand'', ``watch'', ``listen'', \textit{etc.} The actions are exhaustively annotated, which results in 1.58 million action annotations with multiple labels per person. %
The first step of the reorganization is to generate tubelets by linking the labeled bounding boxes at each second. We will link two bounding boxes if they are the consecutive bounding boxes of the same subject with all the action labels being the same. After linking, the tubelets with exactly the same action labels are regarded as semantically corresponding to each other. For example, a tubelet labeled with ``stand + talk to'' semantically corresponds to other tubelets labeled with ``stand + talk to'' as well. The tubelet does not correspond to the tubelets labeled with ``stand'' only, ``talk to'' only, or ``sit + talk to''. It can be understood as that the multiple atomic action labels annotated with one bounding box are combined together. %

Different from spatio-temporal action detection, STVR aims to semantically match video tubelets beyond a predefined category list. Thus, we further split the video tubelets according to their combined labels following~\cite{feng2018video}, so that the training categories have no overlap with the validation or testing categories. We first choose 54 combined labels having over 100 tubelet samples from the 64 validation videos.  27 of them are used for validation and the other 27 are used for testing. After fixing the 54 combined labels, we remove all the frames overlapping with the tubelets belonging to the 54 combined labels in the 235 training videos. The left tubelets in the 235 training videos are used to train our STVR model. The numbers of tubelets belonging to different combined labels are shown in Figure~\ref{fig:hist}.

\begin{figure*}
  \centering
  \includegraphics[width=\textwidth]{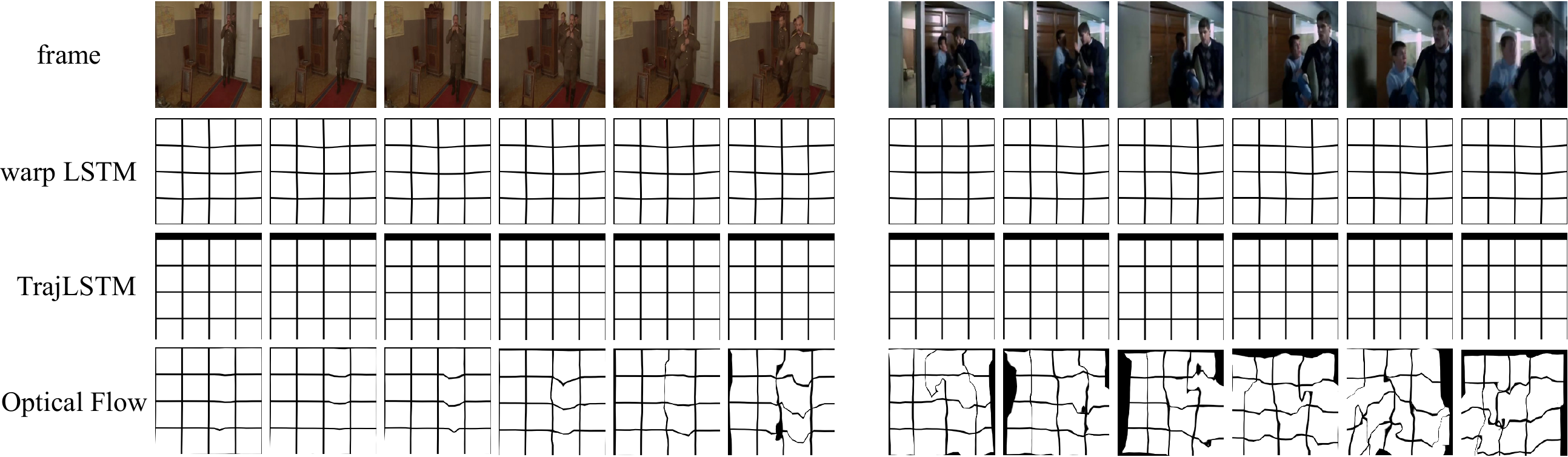}
  \caption{The visualization of the warped grids with different methods. Only one of the five links in TrajLSTM is shown here.}%
  \label{fig:qua_warp}
\vspace{-0.3cm}
\end{figure*}

We describe how to create the query and reference pairs in the following. %
The combined action labels having only one tubelet sample are all discarded because no pair can be formed for this combined label. 
For any query tubelet, we randomly find another tubelet having the same combined label as the target. Then we crop the whole segment containing the target tubelet as the reference video. Such cropping will simplify the STVR task because the temporal boundary is known. To avoid this, we crop a segment longer than the target tubelet so that the reference video contains some background before and after the target tubelet. One thing to mention is that the cropped reference video may contain more than one tubelet having the same label as the query. All of the tubelets in the reference sharing the same label as the query are regarded as target tubelets. During training, the query tubelet and reference video are randomly paired, while the pairs are fixed for validation and testing.%

Following the same intuition, we also reorganize the videos in the UCF-101-24 dataset \cite{soomro2012ucf101} for experiments. Among the 24 action categories, 14, 5, and 5 classes are used for training, validation, and testing, respectively.

\section{Experiments}
We conduct several experiments to verify the effectiveness of warp LSTM in solving the STVR problem. First, three baseline methods are designed and introduced. Then we introduce our experimental settings including evaluation criteria and implementation details. Finally, we report the quantitative results and show the visualizations.

\subsection{Baseline Methods}
Existing spatio-temporal localization methods mainly focus on localizing objects or actions in videos. As far as we know, there is no method specifically designed for STVR. So we design three baseline models for comparison.

\textbf{Clip Independent Baseline.} Clip independent baseline is designed based on the spatio-temporal action localization methods \cite{hou2017tube,kalogeiton2017action}. The reference video is divided into a series of 8-frame clips and the bounding box prediction only depends on the information within the current clip. The clip independent baseline can be implemented by just removing the warp LSTM layer in our proposed model in Figure \ref{fig:framework}.

\textbf{Other ConvLSTM Variants.} The proposed warp LSTM can be viewed as a variant to ConvLSTM \cite{xingjian2015convolutional}. So we create a baseline to compare with the original ConvLSTM by replacing warp LSTM with ConvLSTM. Similarly, we also create a baseline for the comparison with TrajGRU~\cite{shi2017deep}. We replace the polyharmonic interpolation with the structure generating network in \cite{shi2017deep} and name this baseline as TrajLSTM.

\textbf{Optical Flow Baseline.} Warping images by optical flow has been widely used in computer vision research. It is also possible to warp the hidden state of ConvLSTM by the accumulated optical flow. We create another baseline in which the hidden state of ConvLSTM is warped according to optical flow.

\subsection{Experimental Settings}
We resize all the videos to the resolution $320\times320$ before feeding them into the CNN models. The I3D model we use is first initialized by training on the Kinetics dataset \cite{carreira2017quo} and then fine-tuned during the training of our model. To form a batch during the training process, the length of the reference video needs to be fixed. The reference video is fixed to be 2 seconds long by randomly cropping or padding zeros. %
During testing, the query and reference video in full length are fed into the model without batching.
For warp LSTM, we put three horizontal and three vertical lines on the $20\times20$ feature map, which leads to nine control points: \{$(5, 5)$, $(5, 10)$, $(5, 15)$, $(10, 5)$, $(10, 10)$, $(10, 15)$, $(15, 5)$, $(15, 10)$, $(15, 15)$\}. The displacements of the control points are predicted by one CNN layer with a kernel size of $1\times1$. To reduce the number of model parameters, the input-to-state and state-to-state convolutions in warp LSTM are designed following the bottleneck block~\cite{he2016deep}. The $1024$-channel feature map is first projected to $128$-channel and a skip connection is also added from the input to the output of warp LSTM. The following region proposal layers and bounding box regression layers are implemented by Tensorflow Object Detection API \cite{huang2017speed}. The number of links for TrajLSTM is set to be 5. FlowNet2 \cite{ilg2017flownet} is used for optical flow  extraction in the optical flow baseline. The optical flow is resized and rescaled by a factor of $\frac{1}{16}$ to fit the size of the feature map.

\begin{figure*}
  \centering
  \includegraphics[width=\textwidth]{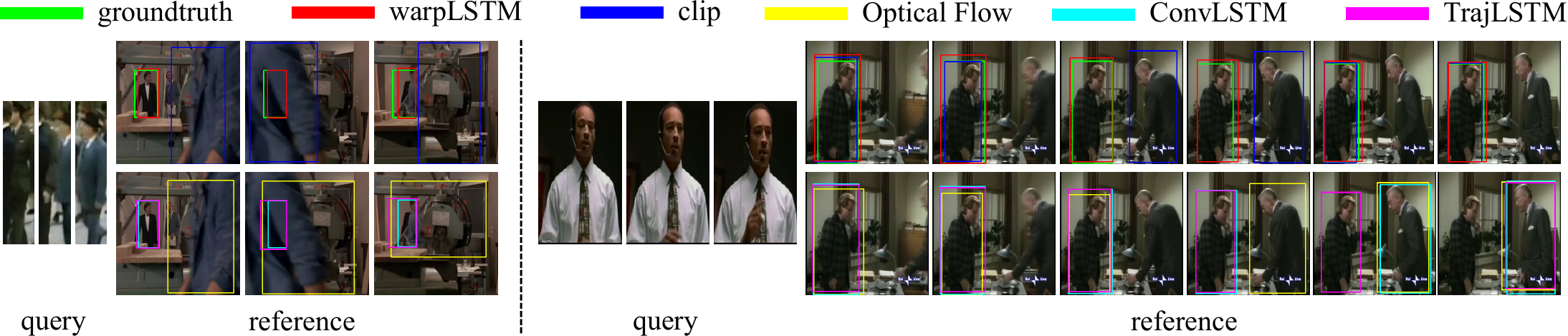}
  \caption{The visualization of the re-localization results. The bounding boxes with the largest confidence of different methods are shown in different colors.}
  \label{fig:qua_box}
\end{figure*}

All the models are trained using stochastic gradient descent (SGD) with momentum value 0.9. The initial learning rate is 0.03 and is divided by 10 after 10k iterations. The batch size we use is set to be 8. It takes about five hours to train one model on four Tesla P40 until the convergence.

\subsection{Evaluation Metrics}
The frame-mAP computed using a modified version of the code\footnote{\url{https://github.com/activitynet/ActivityNet/blob/master/Evaluation/get_ava_performance.py}} released by official the AVA dataset website is reported for evaluation. As described in Sec.~\ref{sec:ava}, there are 27 combined labels for testing. Given a pair of query and reference video, all the labeled bounding boxes in the reference belonging to the same combined label with the query are regarded as ground-truth. The bounding boxes predicted with positive labels are regarded as predictions. The average precision (AP) for one combined label is computed over all the ground-truths and predictions belonging to that combined label with IoU over 0.5. We report the mAP, which is the average of the AP values over the 27 testing combined labels.

\begin{table}
  \caption{The frame-mAP computed with IoU threshold 0.5 of all the methods.}
  \label{tab:results}
  \centering
  \begin{tabular}{c|c|c}
    \toprule
    Method & AVA & UCF-101-24\\
    \midrule
    Clip & 18.8 & 52.9\\
    ConvLSTM \cite{xingjian2015convolutional} & 20.8 & 53.5\\
    Optical Flow & 20.2 & 52.0\\
    TrajLSTM \cite{shi2017deep} & 21.0 & 54.8\\
    Warp LSTM & 21.8 & 59.4\\
    \bottomrule
  \end{tabular}
\end{table}

\subsection{Quantitative Results}
The quantitative results of all the methods are shown in Table \ref{tab:results}. Meanwhile, Figure \ref{fig:qua_warp} shows the warp visualization of two videos in the test split. By comparing the mAP of ``Clip'' and ``'ConvLSTM' baseline, we find that propagating the spatio-temporal information at previous time-steps to current time-step is better than doing the prediction independently for each clip. Using accumulated optical flow to warp the hidden state of the previous time-step leads to worse results than ConvLSTM, which may be because the error is too large in the accumulated optical flow. It can be seen in Figure \ref{fig:qua_warp} that the girds warped by optical flow are noisy. The mAP values of TrajLSTM and ConvLSTM are very similar. The links learned by TrajLSTM on complex action videos seem to be some fixed offsets. The performance of the warp LSTM is the best of all the methods. The results show that propagating the long-term spatio-temporal information by warp LSTM is helpful to STVR.

\subsection{Qualitative Results}
In the second row in Figure \ref{fig:qua_warp}, it can be observed that warp LSTM is able to detect the moving actor and warp the previous feature maps to compensate for the movement. The black in the third and fourth row means that these two methods try to warp some regions outside the feature map into the outputs.
Figure \ref{fig:qua_box} is the visualization of two STVR results. The combined label of the first and second queries are ``walk + talk to + watch'' and ``stand + answer phone'', respectively. In the second clip of the first reference video, the person in the ground-truth bounding box is totally occluded. ``Clip'' and ``Optical flow'' baseline fail to localize correctly because of the occlusion. However, the other three methods are able to handle the short occlusion because they can use the spatio-temporal information in previous clips. In the second example, the two men in the reference video are both standing. The man on the left is labeled with ``stand + answer phone'' and the man on the right is labeled with ``stand + touch + listen to''. It is difficult to distinguish the combined label of these two men because their actions look similar. ``Clip'', ``ConvLSTM'', ``Optical Flow'' and ``TrajLSTM'' make at least one error among the six clips, while the proposed warp LSTM correctly localizes the left man all the time.

\section{Conclusion}
In this paper, we tackled the spatio-temporal video re-localization problem for the first time. Given a query video, spatio-temporal video re-localization aims to find tubelets in a reference video such that the tubelets are semantically corresponding to the given query. Spatio-temporal video re-localization is a natural extension of the temporal video-relocalization \cite{feng2018video} which can be applied to video retrieval and surveillance. To make spatio-temporal video re-localization research possible, we created a new dataset by reorganizing the videos in the AVA dataset~\cite{gu2018ava}. Furthermore, we proposed a matching model to capture the semantic relationship between the query and reference videos. The long-term spatio-temporal information is propagated by a warp LSTM to generate better bounding box predictions. The extensive experimental results show that our proposed method is superior to baseline methods on the spatio-temporal video re-localization task.

In the future, we plan to integrate the warp operation into more sophisticated models such as \cite{girdhar2018better,sun2018actor,zhang2018sentence}.

\section*{Acknowledgement}
This work is partially supported by NSF awards \#1704309,  \#1722847, and \#1813709.

{\small

}

\end{document}